\documentclass[preprint,12pt,num]{elsarticle}
\usepackage{amssymb}
\usepackage{lipsum}
\usepackage{subcaption}
\usepackage{multirow}
\usepackage{booktabs}
\usepackage{amsmath}
\usepackage{makecell}
\usepackage{orcidlink}
\usepackage{eurosym}
\usepackage{bm}

\journal{Next Energy}

\biboptions{sort&compress}

\begin{document}
\begin{frontmatter}

\title{Load Forecasting for Households and Energy Communities: Are Deep Learning Models Worth the Effort?}

\author[inst0]{Lukas Moosbrugger \orcidlink{0009-0005-0447-6529}}
\author[inst0]{Valentin Seiler \orcidlink{0009-0004-7473-336X}}
\author[inst0]{Philipp Wohlgenannt \orcidlink{0009-0008-1723-5451}}
\author[inst3]{Sebastian Hegenbart \orcidlink{0009-0003-2745-1389}}
\author[inst1]{Sashko Ristov \orcidlink{0000-0003-1996-0098}}
\author[inst0]{Elias Eder \orcidlink{0000-0003-3513-6859}}
\author[inst0]{\\Peter Kepplinger \orcidlink{0000-0003-2440-7270} \corref{cor}}
\cortext[cor]{Corresponding author. Email: peter.kepplinger@fhv.at}

\affiliation[inst0]{organization={illwerke vkw Endowed Professorship for Energy Efficiency, Energy Research Centre, Vorarlberg University of Applied Sciences},
            city={Dornbirn},
            country={Austria}}
\affiliation[inst3]{organization={Digital Factory Vorarlberg GmbH},
            city={Dornbirn},
            country={Austria}}
\affiliation[inst1]{organization={Department of Computer Science, University of Innsbruck},
            city={Innsbruck},
            country={Austria}}

\begin{abstract}    

Energy communities (ECs) play a key role in enabling local demand shifting and enhancing self-sufficiency, as energy systems transition toward decentralized structures with high shares of renewable generation. To optimally operate them, accurate short-term load forecasting is essential, particularly for implementing demand-side management strategies. With the recent rise of deep learning methods, data-driven forecasting has gained significant attention, however, it remains insufficiently explored in many practical contexts. Therefore, this study evaluates the effectiveness of state-of-the-art deep learning models---including LSTM, xLSTM, and Transformer architectures---compared to traditional benchmarks such as K-Nearest Neighbors (KNN) and persistence forecasting, across varying community size, historical data availability, and model complexity. Additionally, we assess the benefits of transfer learning using publicly available synthetic load profiles. On average, transfer learning improves the normalized mean absolute error by 1.97 percentage points when only two months of training data are available. Interestingly, for less than six months of training data, simple persistence models outperform deep learning architectures in forecast accuracy. The practical value of improved forecasting is demonstrated using a mixed-integer linear programming optimization for ECs with a shared battery energy storage system. For an energy community with 50 households, the most accurate deep learning model achieves an average reduction in financial energy costs of 8.06\%. Notably, a simple KNN approach achieves average savings of 8.01\%, making it a competitive and robust alternative. All implementations are publicly available to facilitate reproducibility. These findings offer actionable insights for ECs, and they highlight when the additional complexity of deep learning is warranted by performance gains.

\end{abstract}


\end{frontmatter}

\section{Introduction}
\label{introduction}

Energy communities (ECs) are continuously gaining attention, as they accelerate the transition to renewable energy and enhance self-sufficiency by enabling local balancing of supply and demand.
ECs are groups of individuals or organizations that produce, manage, and consume energy locally, through the use of public grid infrastructure, often with the goal of increasing sustainability and reducing energy costs~\cite{european_commission_clean_2019, ponnaganti2023flexibility}. For an efficient operation of these entities---and to improve both self-sufficiency and financial benefit~\cite{vandersteltTechnoeconomicAnalysisHousehold2018}---model predictive control (MPC) methods have been employed with notable success~\cite{mariano2021review, srithapon2023predictive}.

In practice, MPC requires accurate forecasts of exogenous variables such as electrical power demand, heat load, or photovoltaic (PV) production. Many recent studies have been affiliated with heat load forecasting using data-driven or physics-based methods~\cite{hua2024district,powell2014heating,yu2023short}. However, the highest relevance for ECs lies in an accurate prediction of the electric power. A meaningful distinction can be drawn between long-term and short-term power forecasting: the former is crucial for strategic planning and infrastructure development~\cite{royal2025statistical, peterssen2024impact}, whereas the latter is essential for real-time optimization of flexible energy assets~\cite{wazirali2023state, seiler2024assessing}.

Numerous models have been proposed for short-term load forecasting, covering time horizons ranging from one hour up to one week~\cite{wazirali2023state}.
More recently, neural networks using Long Short-Term Memory (LSTM) have been widely regarded as popular models for short-term load forecasting of households~\cite{wazirali2023state,kong_short-term_2019, pallonetto2022forecast}. Within the development of the most competitive deep learning models, many studies have replaced LSTMs with Transformer models due to their ability to handle long-range dependencies and efficiently parallelize computations.

Several recent studies highlight the superior performance of Transformer-based models over LSTMs in the context of short-term load forecasting. For example, Semmelmann et al.~\cite{semmelmann2024impact} compared the forecasting errors of LSTMs and Transformers for an EC consisting of 21 households and concluded that Transformer models outperform LSTMs. Other independently conducted studies~\cite{ran_short-term_2023, zhao_short-term_2021, fang2024method, gao2023adaptive} all came to similar conclusions when evaluating single highly aggregated load profiles representing entire cities and regions. L'Heureux et al.~\cite{lheureux_transformer-based_2022} additionally showed that a Transformer-based model outperforms a state-of-the-art recurrent model when tested on 20 highly aggregated load profiles from an energy utility company. 

In 2024, Beck et al.~\cite{beck2024xlstm} introduced the xLSTM model.
The authors claimed that this model performs favorably when compared to state-of-the-art Transformers, both in performance and scaling. They demonstrated that xLSTM exhibits performance comparable to Transformers, with notable advantages such as linear rather than quadratic growth in memory consumption as the model input size increases. This improvement addresses scalability challenges, making \mbox{xLSTM} a more efficient option to handle larger inputs. A first work by Kraus et al.~\cite{kraus2024xlstm} demonstrates the effectiveness of xLSTM for long-term forecasting across various publicly available time series, with forecast horizons ranging from 96 to 720 hours.  In contrast to prior work, the present study applies xLSTM in a short-term forecasting context, using a 24-hour horizon aligned with the requirements of our MPC application. 

In the context of ECs, the forecast accuracy is often deteriorated by a lack of sufficient historical records~\cite{lee_individualized_2021}, irrespective of the used ML-based model. To address this challenge, transfer learning emerges as a promising solution by enabling knowledge reuse from related data. This approach is based on the idea that knowledge learned during one task can be reused to boost the performance on a related task~\cite{ribeiro_transfer_2018}. In Moosbrugger et al.~\cite{moosbrugger2024improve}, the potential of leveraging transfer learning was demonstrated using open-access synthetic (also called \textit{standard}) load profiles to improve load forecast error for renewable ECs with limited historical data. 
The study utilized a bidirectional LSTM model pre-trained on open-access synthetic profiles, achieving a 62\% reduction in mean squared error (MSE) for forecasted load in an energy community of 74 households, while also significantly improving training stability.

Even though MPC has been extensively applied in the context of ECs, most studies assume either perfect load forecasts or rely on narrowly defined assumptions regarding the forecasting method. These simplifications limit the applicability of the results, as forecasting models are typically tested in isolated scenarios rather than evaluated across a range of conditions such as community size, training data availability, or model complexity, leaving a gap in understanding how different forecasting algorithms perform under varying conditions. This gap is addressed via the following research questions: 

\begin{enumerate}
    \item Is the use of forecasting models based on deep learning methods justified across different EC configurations? 
    \item Can transfer learning using publicly  available synthetic load profiles improve the forecast accuracy of deep learning models in data-scarce ECs?
    \item How is the financial benefit of MPC-based flexibility optimization impacted by the load forecast?
\end{enumerate}

To answer these research questions, load profiles, aggregation levels, training data durations, and model sizes are realistically varied, providing a more comprehensive evaluation. This approach aligns with the recommendation of Pinheiro et al.~\cite{pinheiro_short-term_2023}, who noted that many existing studies demonstrate the superiority of a proposed technique on narrowly defined datasets, often hiding their weaknesses. Therefore, this study provides several contributions: 

\begin{enumerate} 
    \item We quantify the data availability threshold at which advanced deep learning models begin to significantly outperform conventional baseline models.
	\item We propose and evaluate a novel application of transfer learning from publicly  available synthetic load profiles for EC load forecasting.
    \item We demonstrate the financial implications of forecast accuracy through a case study of a simulated EC, highlighting the economic benefits of using simple models such as k-nearest neighbors (KNN).
\end{enumerate}

This study therefore critically assesses the suitability of current deep learning approaches for household and EC load forecasting by systematically probing their sensitivity to real-world scenarios and investigating the value of improved forecasting accuracy for energy system optimization.

\section{Methods}
\label{sec:methods}

\begin{figure}[h]
	\centering
	\includegraphics[width=0.6\columnwidth]{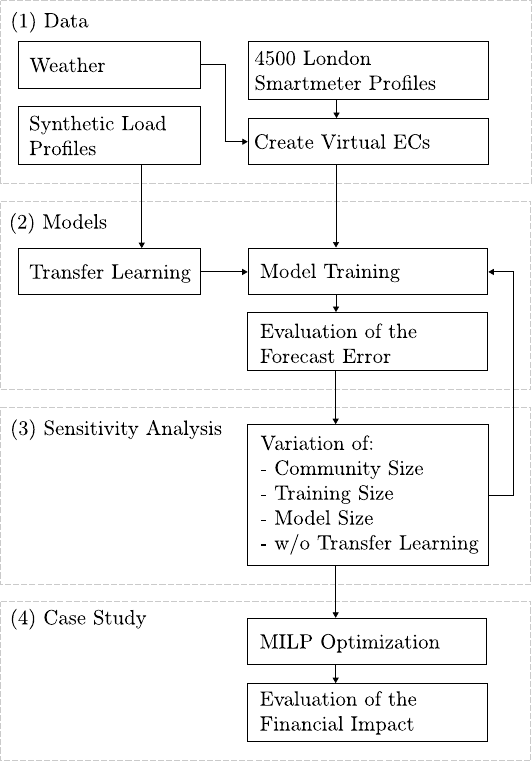}
    \caption{Summary of the research design for the present study.}
	\label{Figure_1}
\end{figure}

To assess the impact of load forecast accuracy on potential financial benefits of storage optimization, a comprehensive framework was developed. This framework, and the data used, are publicly available in a GitHub repository~\cite{github_loadforecasting2024}. It includes: (1) a performance benchmark of data-driven forecasting algorithms, (2) a sensitivity analysis of the models developed, and (3) an evaluation of their individual financial impacts within a case study. Fig.~\ref{Figure_1} provides an overview of the methodological approach used in this study.

In accordance with Fig.~\ref{Figure_1}, different types of data were used and curated to evaluate various models regarding their load forecasting capabilities. Typical input features for short-term load forecasting were derived, including datetime information, weather conditions, and historical load data, which corresponds to other load forecasting studies~\cite{pinheiro_short-term_2023, royal2025statistical, semmelmann2024impact,powell2014heating, gao2023adaptive, chen2021theory}.

In the next step, three deep learning models (xLSTM, LSTM, and Transformer) were compared against two baseline models: Persistence Prediction and \mbox{k-Nearest} Neighbors (KNN). Each model performs day-ahead short-term load forecasting, predicting an hourly power load profile from midnight to the following midnight. The models were trained and tested on various configurations in terms of community size, training dataset size, and model complexity. The practical value of improved forecasting was analyzed through a mixed-integer linear programming (MILP) optimization model for ECs with shared battery energy storage. 

\subsection{Data description}
\label{sec_Dataset}

This study is based on 4,500 household load profiles from the UK Power Networks smart meter dataset~\cite{uk_power_networks_2015}, which contains electricity consumption data from households in London. To construct virtual ECs, subsets of 1, 2, 10, 50, or 100 households were randomly selected from the dataset. This selection process was repeated 20 times for each community size, without replacement, resulting in $100$ unique virtual ECs.
All profiles span the period from July 2012 to February 2014. Q4 of 2013 was used as the test period, with the preceding 12 months used for training in all baseline model evaluations. The effect of using different quarters of 2013 as the testing period was also evaluated in a robustness analysis, which can be found in \ref{AppendixB}. 

In addition to the London smart meter data, synthetic load profiles were used to pretrain the deep learning models, as detailed in Section~\ref{sec_Models}.
For these purposes, the German synthetic load profiles provided by the Python package \textit{demandlib} \cite{ceruti2016demandlib} were employed. This package generates hourly synthetic load data for an arbitrary year and can even forecast loads for future years, making it a useful resource for transfer learning applications.

Besides the above load profiles, all available Meteostat \cite{Lamprecht_Meteostat_Python} weather features for London from 2012 to 2014 were utilized. These features include ambient temperature, dew point, wind direction, wind speed, air pressure, and relative humidity.

\subsection{Machine Learning Methods}
\label{sec_Models}

For each of the virtual ECs, two baseline models and three deep learning models were trained and evaluated. A detailed description of the input features and the model architecture is provided in this section.

\subsubsection{Input Features}

The current datetime, past weather data, and past load values were used as model input features, consistent with previous load forecasting studies~\cite{pinheiro_short-term_2023, royal2025statistical, semmelmann2024impact,powell2014heating}. This resulted in 20 input features in total, which are detailed as follows:

\begin{itemize}
    \item \textbf{Day of the week}: Encoded as a one-hot vector, represented as a binary 7-dimensional vector with one "hot" bit and public holidays assigned to Sunday~\cite{semmelmann2024impact, ziel2018modeling}.
    \item \textbf{Hour of the day}: Encoded cyclically. The hour of day was mapped onto a unit circle using sine and cosine transformations to preserve its cyclical nature.
    \item \textbf{Day of the year}: Encoded cyclically, similar to the hour of day feature.
    \item \textbf{Lagged load profiles}: These features include the hourly load measurements from one, two, and three weeks prior to the current timestep. 
    Preliminary analysis showed that these days exhibit the highest correlation with the target values, outperforming more recent lags such as the previous day.    
    \item \textbf{Weather data}: All weather features described in Section~\ref{sec_Dataset} from the 24 hours preceding the forecast were used as input features for the model.
\end{itemize}

\subsubsection{Deep Learning Methods}
\label{DeepLearningModels}

Two widely used deep learning methods for load forecasting—LSTMs and Transformers—were selected for this comparative study, based on findings from the literature review \cite{wazirali2023state,kong_short-term_2019,pallonetto2022forecast,semmelmann2024impact,ran_short-term_2023,zhao_short-term_2021,lheureux_transformer-based_2022,fang2024method}. 
Additionally, the recently proposed xLSTM model~\cite{beck2024xlstm} was included in the evaluation.

\begin{figure}[htbp]
	\centering
	\includegraphics[width=\columnwidth]{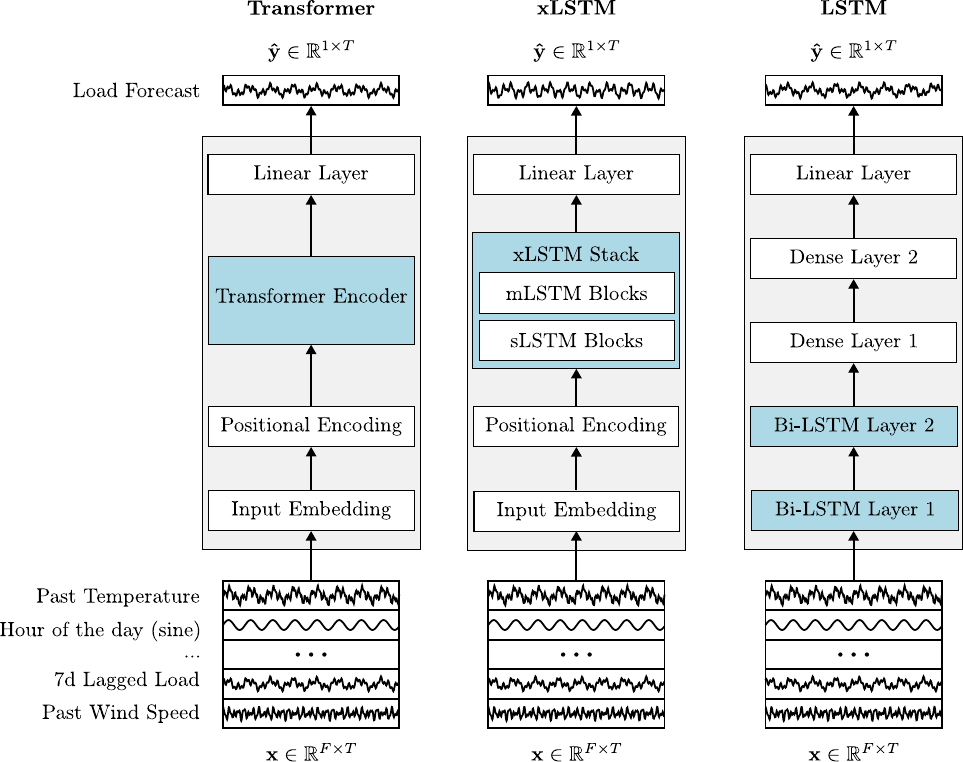}
	\caption{
    Overview of the sequence-to-sequence deep learning models evaluated in this study. Here, $T$ represents the number of timesteps (24 hours), and $F$ denotes the number of features (20).
    }
	\label{Figure_2}
\end{figure}

Figure~\ref{Figure_2} illustrates the deep learning architectures used. We adopted the established LSTM-based architecture used in Kong et al.~\cite{kong_short-term_2019} by using multiple dense, i.e., fully connected layers as a single regression head for predicting the load. An encoder-only Transformer architecture was employed following Nie et al.~\cite{nie2022time}. For xLSTMs, we adopted the approach and the parameter settings provided in the reference implementation by Beck et al.~\cite{beck2024xlstm}.

For the sake of a fair comparative study, the model sizes, i.e., the number of trainable parameters, were set to be comparable across all architectures. Empirical results indicate that 5k parameters serve as a robust and fair baseline for this problem setting. 
A detailed summary of all model parameters used can be found in~\ref{AppendixC}.

All models were optimized using Adam \cite{KingmaB14} across 100 training epochs, with a batch size of 256 and a learning rate schedule that declined in four stages: 0.01, 0.005, 0.001, and 0.0005.

In a previous study~\cite{moosbrugger2024improve}, we already applied transfer learning by using synthetic load profiles. This approach was extended to the present study. During pre-training, models were exposed to openly available synthetic load profiles (as described in Section~\ref{sec_Dataset}) spanning the entire period from July 2012 to February 2014. The learned parameters from this phase were then used in a fine-tuning step of the model using the target EC profiles.

The models were trained using the mean absolute error (MAE) loss, as energy applications typically prioritize minimizing absolute errors rather than squared errors, due to the mostly linear relation between energy costs and used energy:

\begin{equation}
\text{MAE} = \frac{1}{n} \sum_{i=1}^n \left| y_i - \hat{y}_i \right|
\label{eq:mae}
\end{equation}

To enable comparisons across ECs of varying sizes within the same table or figure, the results were evaluated using the normalized mean absolute error (nMAE):

\begin{equation}
\text{nMAE} = \frac{\text{MAE}}{\bar{y}}
\label{eq:nmae}
\end{equation}

where the normalization factor \(\bar{y}\) represents the arithmetic mean.

The individual models were trained on an Ubuntu 22.04 system equipped with 32 CPU cores running at 3.2 GHz and 132 GB of RAM.
To ensure reproducibility, a conda environment file is supplied, detailing the primary dependencies, including Python 3.11 \cite{python311} and PyTorch 2.2 \cite{pytorch2023}.

\subsubsection{Benchmark Methods}

The aforementioned state-of-the-art deep learning methods were benchmarked against two simple and commonly used methods:

\begin{itemize}
    \item Persistent Prediction Method: This benchmark method predicts the load of the next day to be exactly the same load as it was seven days ago. Gross et al. \cite{gross2021comparison} call this the naive seasonal model.  
    \item KNN~\cite{gross2021comparison}: A nearest neighbor regressor, using the same input features as the deep learning methods (see Section~\ref{sec_Dataset}). The specific KNN configuration was selected based on preliminary experiments on the training set. The number of neighbors was set to $k=40$, meaning that each hourly load value is predicted based on the 40 most similar historical hours, determined using Euclidean distance.  Additionally, closer neighbors are assigned greater influence through distance-based weighting. 
\end{itemize}

\subsection{Sensitivity Analysis}
\label{sec:SensitivityAnalysis}
\begin{figure}[htbp]
	\centering 
	\includegraphics[width=0.5\columnwidth]{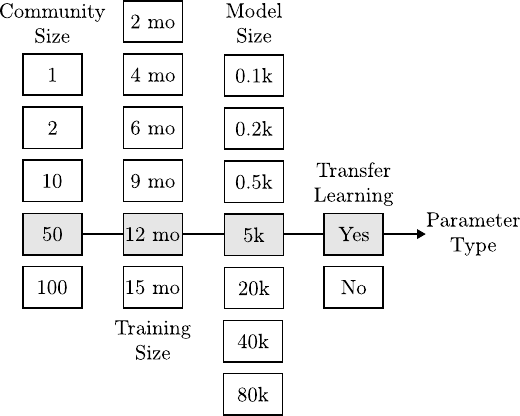}	
    \caption{Forecast error sensitivity analysis. Each dimension is varied independently, with all other parameters held at their baseline, as indicated by the grey boxes above.}
    \label{Figure_3}
\end{figure}

\begin{figure}[htbp]
	\centering
	\includegraphics[width=0.5\columnwidth]{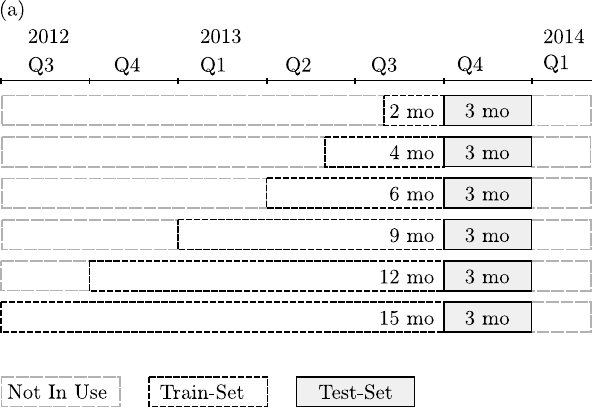}
        \caption{Data split strategy: The final quarter of 2013 (Q4) serves as the testset, with varying lengths of the immediately preceding data used for training. Test results for other seasonal periods are provided in the \ref{AppendixA}.}
        \label{Figure_4}
\end{figure}

The sensitivity of the models was analyzed with respect to varying EC properties, levels of data availability, and model properties.
The experiments were specifically designed to evaluate the impact of the community size, training size, model size and the effectiveness of transfer learning. The approach is illustrated in Fig.~\ref{Figure_3}. The baseline configuration (depicted in grey), as explained in the individual sections above, is defined with the following parameters: A model size of 5k parameters, a community size of 50 households, and a training set spanning 12 months. From this baseline, a systematic exploration was conducted to examine how the forecast errors depend on the individual parameters, which are subsequently explained in greater detail: 

\begin{itemize}
    \item \textbf{Community size:} To investigate different levels of aggregation, five different community sizes were selected for comparison. 
    According to the Austrian Coordination Office for Energy Communities~\cite{austrian_energy_communities}, most ECs in Austria fall into one of three size categories: 2–10, 10–50, or 50–100 households.
    Based on this classification, the boundary cases of 1, 2, 10, 50, and 100 households were chosen for this study. 
    In the parameter grid, 50 households represent the baseline community size.
    \item \textbf{Training size:} We further considered a variety of training dataset sizes, to evaluate the practical challenge of insufficient training data. The amount of available training data was set to 2 months for the first evaluation, then progressively increased to 4, 6, and 9 months. To ensure that each trained model has been exposed to all seasons of a year, a reasonable training period is comprised of 12 months. We therefore used 12 months as a robust baseline. Furthermore, 15 months of data were used, representing the maximum amount of available training data in the dataset used for the investigation.  The sequential data split strategy is illustrated in Figure~\ref{Figure_4}.
    \item \textbf{Model size:} To assess the impact of the model size, each model was constructed with a different number of parameters—targeting approximately 0.1k, 0.2k, 0.5k, 5k, 20k, 40k, and 80k.
    \item \textbf{Transfer learning:} To quantify the impact of transfer learning, model performance was evaluated across different training set sizes, both with and without applying transfer learning.
\end{itemize}

\subsection{Case Study on the Financial Implications of Load Forecasting}

\begin{figure}[h]
	\centering 
	\includegraphics[width=0.5\columnwidth]{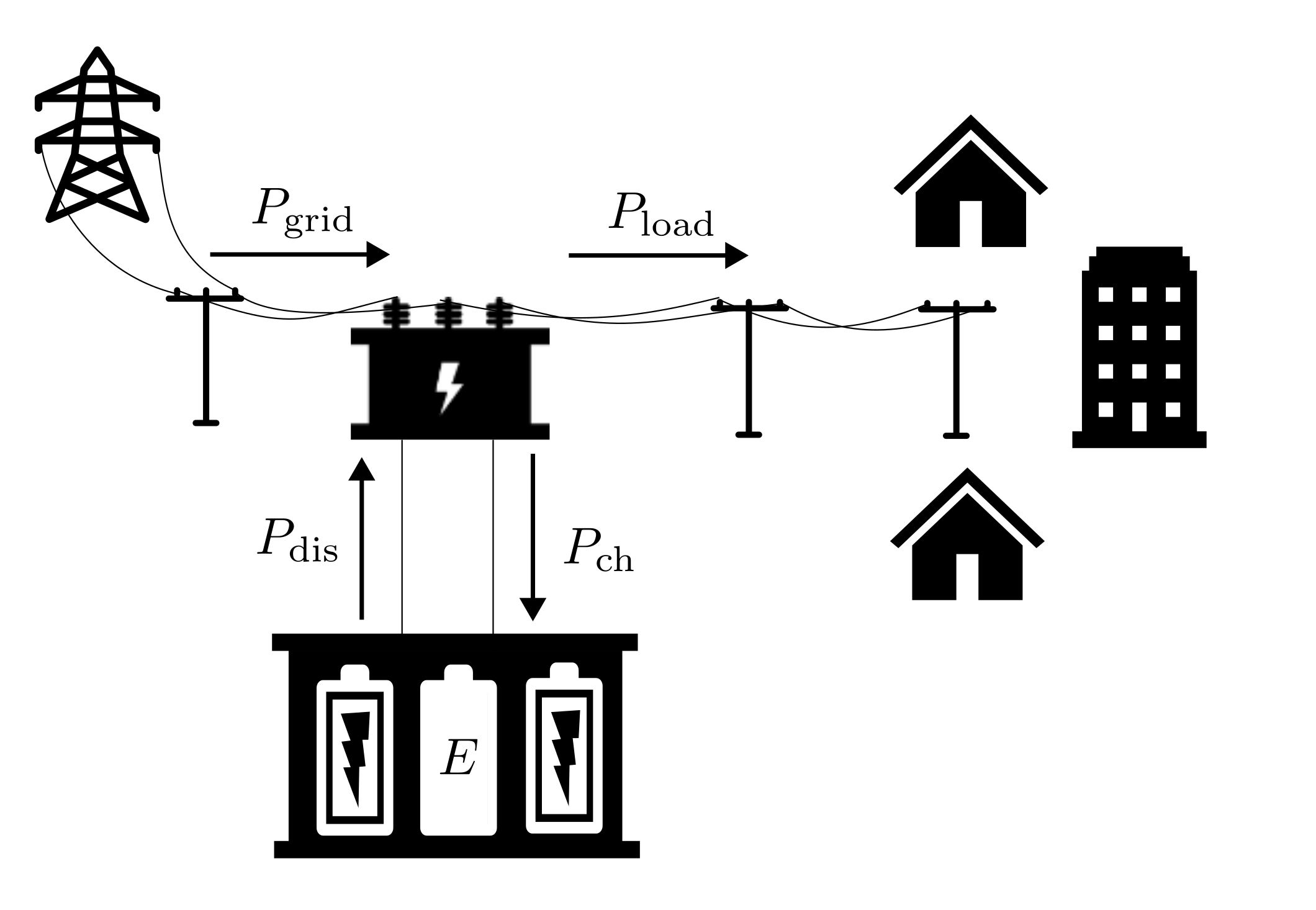}	
	\caption{Scheme of the case study EC with community BESS. $E$ denotes the energy stored in the community BESS, $P_{\text{ch}}$ is the charging power, $P_{\text{dis}}$ is the discharging power, $P_{\text{load}}$ is the electrical load of the EC and $P_{\text{grid}}$ is the consumed power from the grid, which has to be purchased.} 
	\label{Figure_5}
\end{figure}

Using the following case study, the financial impact of load forecasting errors in ECs was evaluated by utilizing a community battery energy storage system (BESS). The general goal was to reduce daily energy costs via intraday load shifting under a real-time price (RTP) tariff. The BESS was controlled through MPC, utilizing the forecasting models above for the power load. Details of the case study design are subsequently described. Fig.~\ref{Figure_5} gives a schematic overview of the case study.

The EC was modeled with a RTP based on the Austrian tariff structure. The price index of the Energy Exchange Austria (EXAA) \cite{case_study_exaa} was used in combination with network usage fees and taxes according to current Austrian standards to form the EC tariff. To evaluate optimal battery size, two schemes were applied: 
\begin{enumerate}
    \item A per-household-capacity of $12~\mathrm{kWh}$ was assumed---which is the optimal capacity per household according to Barbour et al.~\cite{barbour2018community}---and scaled linearly with community size. 
    \item A fixed community size of 10 households was used to evaluate the optimal BESS capacity in a range of 5 kWh to 200 kWh, aligning with typical community-scale battery systems noted in~\cite{barbour2018community}. 
\end{enumerate}

A 4-hour (C-Rate = 0.25) BESS with a maximum power of $P_\text{max}$ is installed with symmetric charging and discharging efficiencies of 
$\eta_{\text{ch}}=\eta_{\text{dis}} = 92.2\%$, respectively, resulting in a round-trip efficiency of 85\%, which is in accordance with \cite{cole_cost_2021}. The simulation was executed for a test period ranging from October 1, 2013, to December 31, 2013, with the time period discretized to an hourly resolution. The optimization was conducted once per day to determine the optimal future schedule.


The optimization model distinguishes between state and process variables. State variables are indexed via $t$ and defined for one point in time, while process variables are indexed via $p$ and defined for one time period. $N$ periods result in $N+1$ time points. $E_t$ is the energy content of the BESS at time point $t$ and $P_{\text{ch}, p}$, $P_{\text{dis}, p}$, $\hat{P}_{\text{load}, p}$, $P_{\text{grid}, p}$ are the charging, discharging, load and grid powers for each period $p$, respectively. The MILP uses the load forecast $\hat{P}_{\text{load}, p}$ and the RTPs $\pi_p$ as inputs. The resulting daily optimization problem can be formulated as follows:
\begin{align}
    \label{casestudy_objective}
    &\min_{{P}_{\mathrm{ch}}, P_{\mathrm{dis}}} \sum_{p \in \mathcal{P}} P_{\text{grid},p} \pi_p \Delta t\\
    \label{eqz1b}
    &\mathcal{P} = \left\{0, \ldots, 23\right\}\\
    \label{eqz1c}
    &\mathcal{T} = \left\{0, \ldots, 24\right\}\\
    \text{s.t. }\forall p \in \mathcal{P}: \nonumber\\
    \label{casestudy_grid_balance}
    \quad &P_{\text{grid}, p} + P_{\text{dis}, p} = P_{\text{ch}, p} + \hat{P}_{\text{load}, p}\\
    \label{casestudy_min_grid_power}
    \quad &P_{\text{grid}, p} \geq 0.15 \hat{P}_{\text{load}, p}\\
    \label{casestudy_battery_balance}
    \quad &E_{p+1} - E_{p} = \left(\eta_{\text{ch}}P_{\text{ch}, p} - \frac{P_{\text{dis}, p}}{\eta_{\text{dis}}} \right) \Delta t\\
    \label{casestudy_bool1}
    \quad & 0 \leq P_{\text{ch}, p} \leq b_{\text{ch}, p} P_\text{max}\\
    \label{casestudy_bool2}
    \quad & 0 \leq P_{\text{dis}, p} \leq b_{\text{dis}, p} P_\text{max}\\
    \label{casestudy_bool3}
    \quad &b_{\text{ch}, p} + b_{\text{dis}, p} = 1\\
    \text{and s.t. } \forall t \in \mathcal{T}: \nonumber\\
    \label{casestudy_min_max_energy}
    \quad &E_{\text{min}} \leq E_{t} \leq E_{\text{max}}\\
    \text{and s.t.:} \nonumber\\
    \label{casestudy_start_e}
    &E_{0} = E_{\text{start}}\\
    \label{casestudy_end_e}
    &E_{24} = E_{\text{end}}\\
    \text{with:} \nonumber\\
    \quad & E=\left(E_0,\cdots, E_{24}\right) \in \mathbb{R}^{25}\\
    \quad & P_{\text{ch}}=\left(P_{\text{ch}, 0}, \cdots, P_{\text{ch}, 23} \right) \in  \mathbb{R}^{24}\\    
    \quad & P_{\text{dis}}=\left(P_{\text{dis}, 0}, \cdots, P_{\text{dis}, 23} \right) \in  \mathbb{R}^{24}\\    
    \quad & P_{\text{grid}}=\left(P_{\text{grid}, 0}, \cdots, P_{\text{grid}, 23} \right) \in  \mathbb{R}^{24}\\    
    \quad & \hat P_{\text{load}}=\left(\hat P_{\text{load}, 0}, \cdots, \hat P_{\text{load}, 23} \right) \in  \mathbb{R}^{24}\\  
    \quad & b_{\text{ch}}=\left(b_{\text{ch},0},\cdots,b_{\text{ch},23}\right) \in \{0, 1\}^{24}\\
    \quad & b_{\text{dis}}=\left(b_{\text{dis},0},\cdots,b_{\text{dis},23}\right) \in \{0, 1\}^{24}
\end{align}
Eq.~\ref{casestudy_objective} defines the objective function minimizing the daily electrical energy costs. Eqs.~\ref{eqz1b} and~\ref{eqz1c} define sets of indices for all time periods and time points, respectively.
The power balance of the EC is defined in Eq.~\ref{casestudy_grid_balance}, while Eq.~\ref{casestudy_min_grid_power} forces the solution of the optimization problem to fulfill that the grid supplies at least 15\% of the total electrical load, based on the forecast. This constraint acts as a buffer to prevent feed-in of excess energy into the power grid, which can occur if the actual load is significantly lower than the forecasted load—causing the BESS to discharge energy that is not consumed. Since feed-in is not compensated under standard electricity tariffs, such scenarios should be avoided. Note that Eq.~\ref{casestudy_min_grid_power} is not contained in the perfect prediction scenario.
Eqs.~\ref{casestudy_battery_balance}-\ref{casestudy_min_max_energy} describe the BESS model, where $\eta_{\text{ch}}$ and $\eta_{\text{dis}}$ are the charging and discharging efficiencies, respectively. $E_{\text{min}}$ and $E_{\text{max}}$ denote minimum and maximum BESS capacity, respectively. The energy balance of the BESS is described in Eq.~\ref{casestudy_battery_balance}.
Eqs.~\ref{casestudy_bool1}-\ref{casestudy_bool3} prevent simultaneous charging and discharging via the binary variables $b_{\text{ch}}$ and $b_{\text{dis}}$ and to enforce the maximum power limitation.
The initial and end conditions of the BESS are defined by Eqs.~\ref{casestudy_start_e} and \ref{casestudy_end_e}.
Note that the optimization uses the forecasted load $\hat{P}_{\text{load}}$ without a feedback loop, as real-time measurements of up to 100 households can be considered very challenging in practice. Consequently, the grid is used to compensate for forecast errors. 
The resulting grid power $P_{\text{grid},p}$ at period $p$, employing the solution of the optimization problem to the realized EC load, can therefore be calculated by 
\begin{align}
    P_{\text{grid},p} &= P_{\text{ch},p} - P_{\text{dis},p} + P_{\text{load},p}.
\end{align}

Here $P_{\text{load},p}$ is the resulting load of the EC in period $p$.

\section{Results and Discussion}
\label{sec:results}

In the following section, key findings are presented, whereas detailed numerical results, including comprehensive results for all models, are provided in \ref{AppendixA}. 

\subsection{Preliminary Analysis}

The preliminary analysis aims to provide context and support for the core findings presented in later sections.

\vspace{\baselineskip}
\noindent \textbf{Linear Feature Correlation}

\begin{figure}[h]
    \centering
    \includegraphics[width=\textwidth]{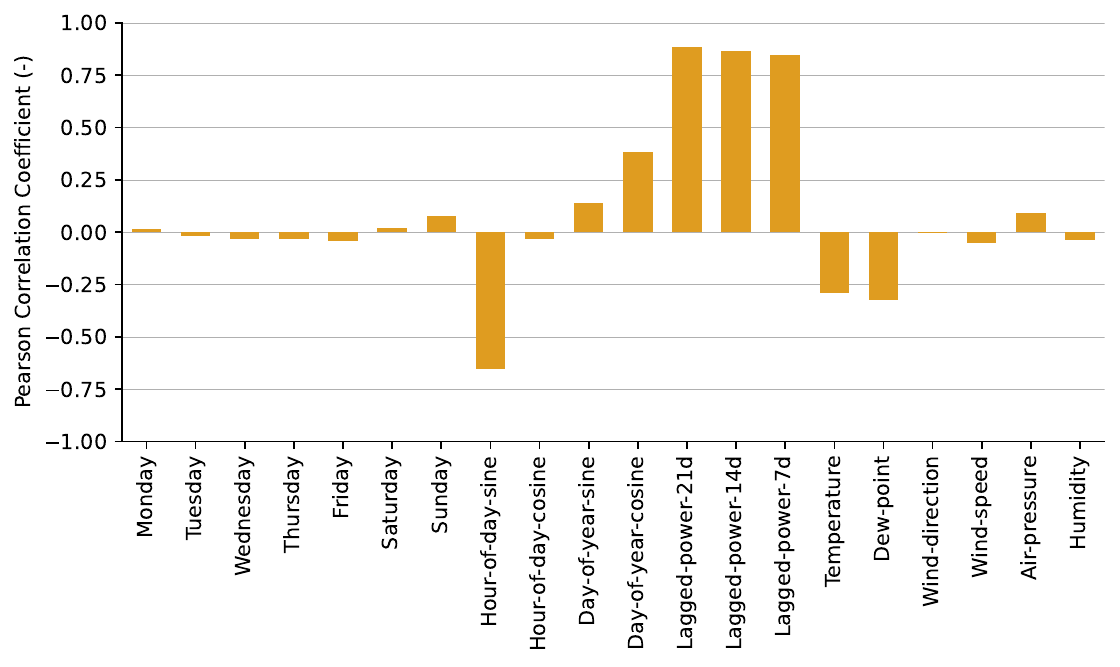}
    \caption{Linear correlation between each input feature and the target load profile.   
    }
    \label{Figure_6}
\end{figure}

Fig.~\ref{Figure_6} illustrates the Pearson correlation coefficient among individual features within an EC comprising 50 households over a 20-month period. 
It confirms that, as expected, features such as lagged power values and day-of-year exhibit significant correlations with the target load power profile. 
On the other hand, features such as ambient temperature and dew point are negatively correlated with load power, as lower temperatures are associated with increased heating and lighting demand throughout the year.
The sine part of the sinusoidal hour-of-day encoding is also strongly negatively correlated, since it reaches its minimum around 6 p.m., which is typically aligned with the daily maximum in load demand. The strong correlation with lagged load values represents the reasoning for using a persistence model for short-term forecasting.

\vspace{\baselineskip}
\noindent \textbf{Forecast for One Sample Energy Community}

\begin{figure}[h]
    \centering
	\includegraphics[width=\textwidth]{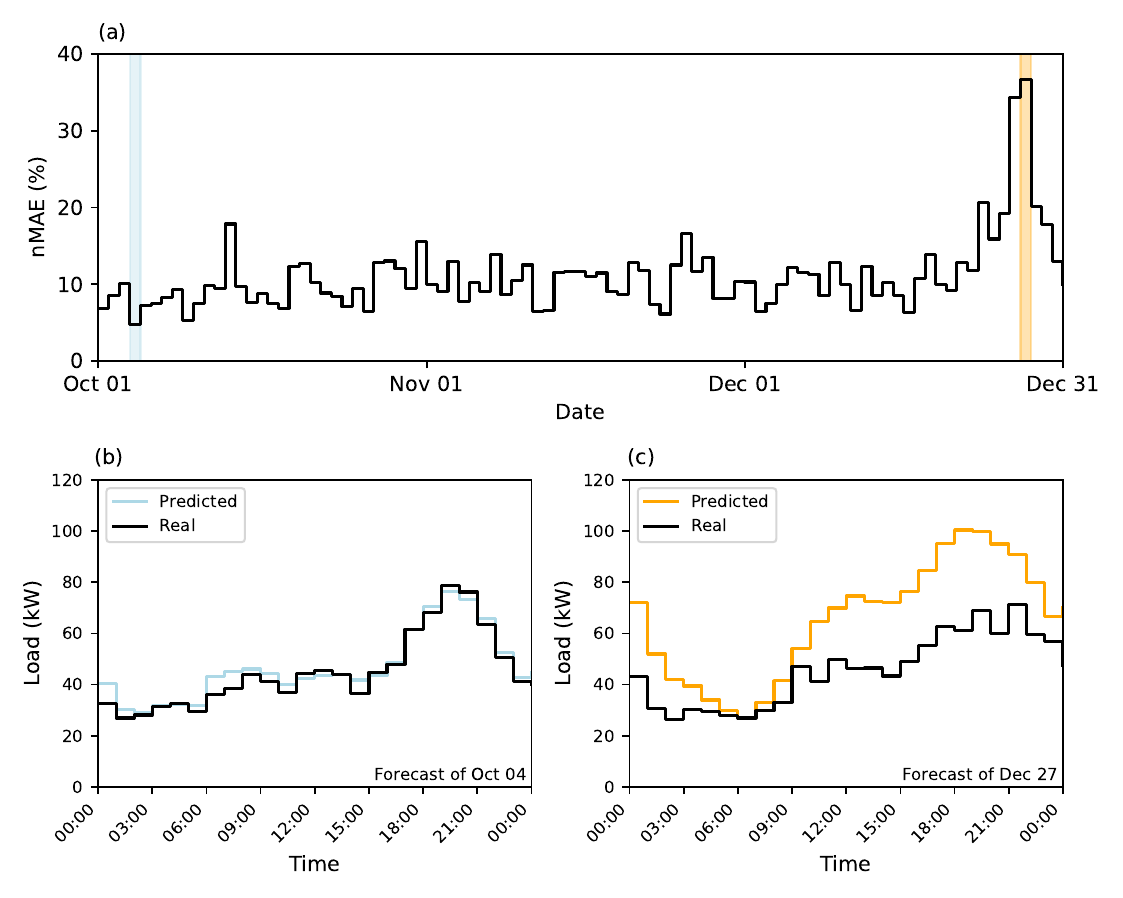}
    \caption{Performance of the Transformer model on a sample EC using the baseline configuration. 
    (a)~Daily forecast errors across the entire test period. 
    (b)~Example of a day with a relatively low forecast error (nMAE = $4.8\%$).
    (c)~Example of a day with a relatively high forecast error (nMAE = $36.6\%$), where the entire period from noon to midnight is clearly overestimated.
    }
    \label{Figure_7}
\end{figure}

Fig.~\ref{Figure_7} shows daily forecasts for a sample EC under the baseline configuration.
It indicates that the forecast error remains low for the majority of the test set.
In particular, the daily nMAE from October to mid-December fluctuates around 10\% on most days.
An example load profile of a day with strong forecast accuracy is shown in Fig.~\ref{Figure_7}(b).
As expected, forecast errors peak during the Christmas holidays, reflecting the distinct energy consumption patterns of the holiday season. 
This anomaly is illustrated in Fig.~\ref{Figure_7}(c).

\subsection{Sensitivity Analysis}

This subsection presents how changes to key parameters affect the load forecasting error.

\vspace{\baselineskip}
\noindent \textbf{Training with Limited Data}

\begin{figure}[h]
	\centering 
	\includegraphics[width=\columnwidth]{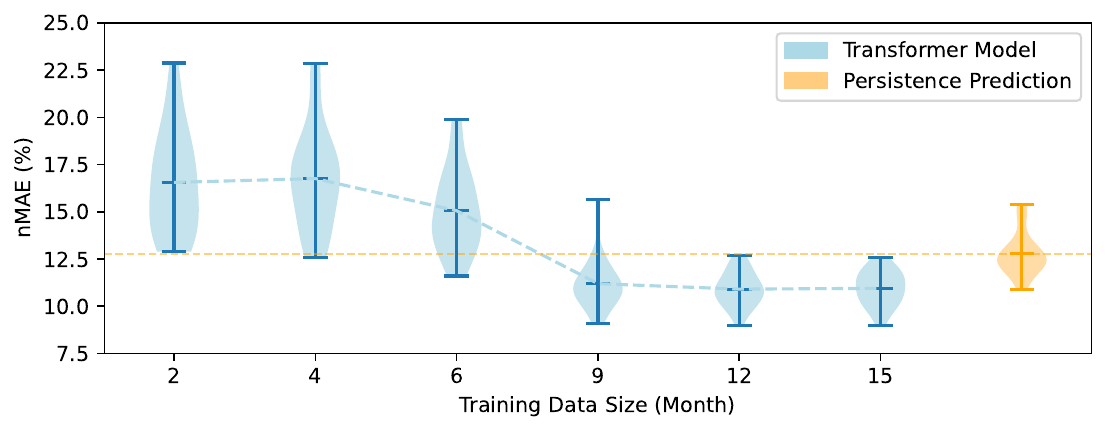}	
	\caption{Minimum, arithmetic mean, maximum, and distribution of load forecasts with different sizes of training data. Each setting includes results from 20 ECs, based on the baseline configuration. While results for the Transformer are presented, the behavior is the same for other deep learning models and KNN, as shown  in \ref{AppendixA}.
    } 
	\label{Figure_8}
\end{figure}

Fig.~\ref{Figure_8} illustrates the performance improvement of the deep learning models compared to the persistence prediction benchmark, both regarding the available training data. Notably, the persistence prediction is independent of the available amount of training data. 

It is seen that the nMAE of the persistent predictor is smaller compared to the deep learning models with up to six months of training data. 
However, with nine months of training data or more, the deep learning models outperform persistent prediction, with a difference in nMAE ranging from 1.8 percentage points to 2.0 percentage points, depending on the model type. This finding is particularly valuable for newly established ECs, which can initially rely on persistence prediction for MPC and transition to more sophisticated models after operating for nine months.

Fig.~\ref{Figure_8} further demonstrates that the forecast error decreases significantly as the size of the training data increases. The models demonstrate improved generalization to new days when trained on a larger volume of historical data. However, this reduction in error appears to plateau at approximately 12 months of training data, indicating diminishing returns from additional historical data beyond this point. This suggests that the model effectively captures all temporal patterns within that training period.

\begin{figure*}[h]
	\centering
	\includegraphics[width=\textwidth]{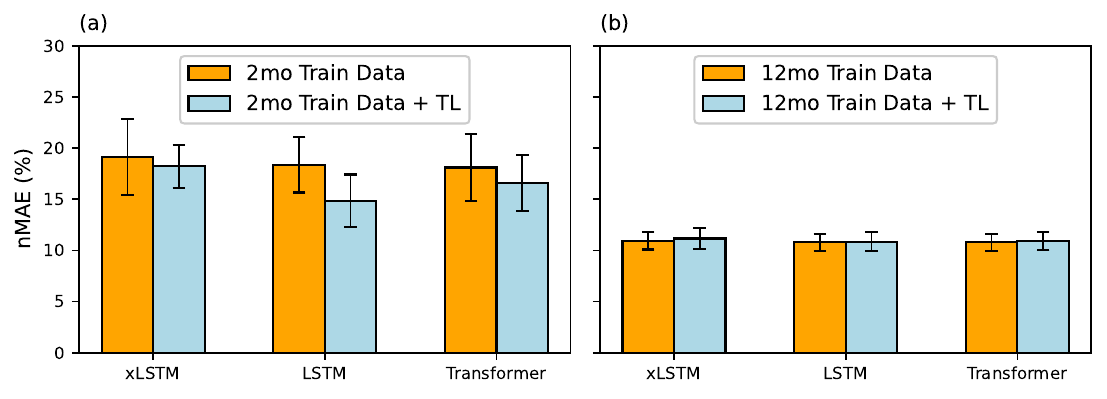}
    \caption{Forecast error of deep learning models with and without transfer learning (TL), using (a) a training set of 2 months and (b) a training set of 12 months. The column shows the mean over 20 ECs. All other settings follow the baseline configuration.}
	\label{Figure_9}
\end{figure*}

Fig.~\ref{Figure_9} depicts the nMAE of the investigated deep learning models trained with and without transfer learning. The varying effect of transfer learning is emphasized by providing a comparison of models trained with two months of available training data (a), and models trained with 12 months of available training data (b). 

As illustrated in Fig.~\ref{Figure_9}(a), transfer learning using synthetic load profiles shows a notable reduction in nMAE for two months of available training data. 
Across all trained models, the average nMAE improvement is 1.97 percentage points. 
Remarkably, this improvement is achieved despite using synthetic profiles from Germany rather than the UK for pretraining. Fig.~\ref{Figure_9}(b) illustrates that as training duration increases, the benefits of transfer learning diminish---and may even become slightly negative. This is because the model eventually learns all relevant temporal patterns from the target data itself, leaving little advantage to be gained from the pretrained knowledge. 

This finding underscores the versatility and generalizing ability of synthetic profiles for pretraining, for cases with limited training data. Unlike specific profiles that may risk overfitting the models, synthetic profiles demonstrate broad applicability across diverse scenarios.

\vspace{\baselineskip}
\noindent \textbf{Impact of the Model Size}
\label{sec_impact_of_model_size}

\begin{figure}[h]
    \centering 
    \includegraphics[width=0.5\columnwidth]{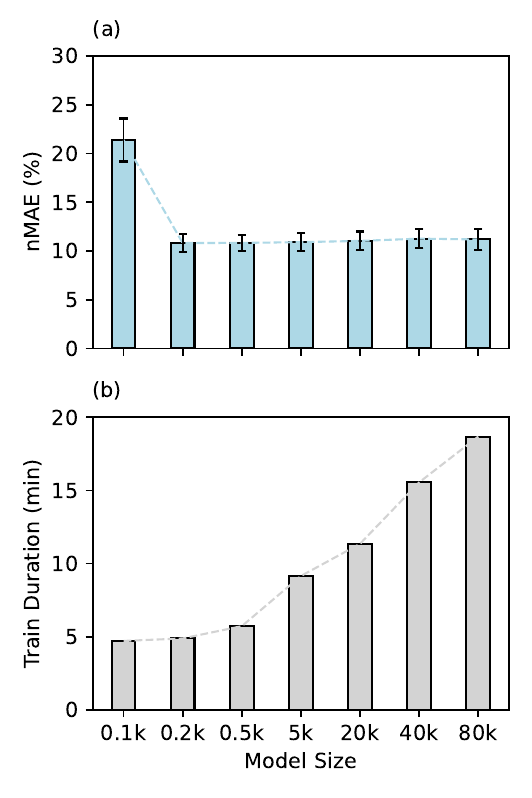}	
    \caption{
    Load forecasting nMAE illustrated using the Transformer model (a), and total training duration of the entire forecasting framework (b). All computations are based on results from 20 ECs under the baseline configuration. The whiskers in (a) indicate the standard deviation over all ECs.
    }
    \label{Figure_10}
\end{figure}

In Fig.~\ref{Figure_10}, the impact of the model complexity is illustrated, thus nMAE (a) and training duration (b) are depicted for varying model sizes. These results reveal several noteworthy insights.

The Transformer model's accuracy is very stable across varying model sizes, i.e., across different hyperparameter settings. For model sizes of 0.2k parameters or more, the average nMAE fluctuates at most by 0.45 percentage points, while the standard deviation remains nearly independent of the model's parameter count, with a maximum fluctuation of 0.27 percentage points. 

This illustrates that higher-capacity models, due to the larger number of parameters, do not necessarily result in higher accuracy. Quite opposing, the models can be configured to have as few as 200 parameters, yet still deliver a sufficiently low nMAE. This finding is especially relevant for small edge computing devices, such as home automation systems with limited computational resources, as it suggests that smaller models can achieve comparable performance without the need for extensive computational power. As Fig.~\ref{Figure_10}(b) illustrates, the increase in training duration, although moderate, is not justified by a corresponding decrease in nMAE. 

\vspace{\baselineskip}
\noindent \textbf{Impact of Different Community Sizes}

\begin{figure}[h]
	\centering 
	\includegraphics[width=\columnwidth]{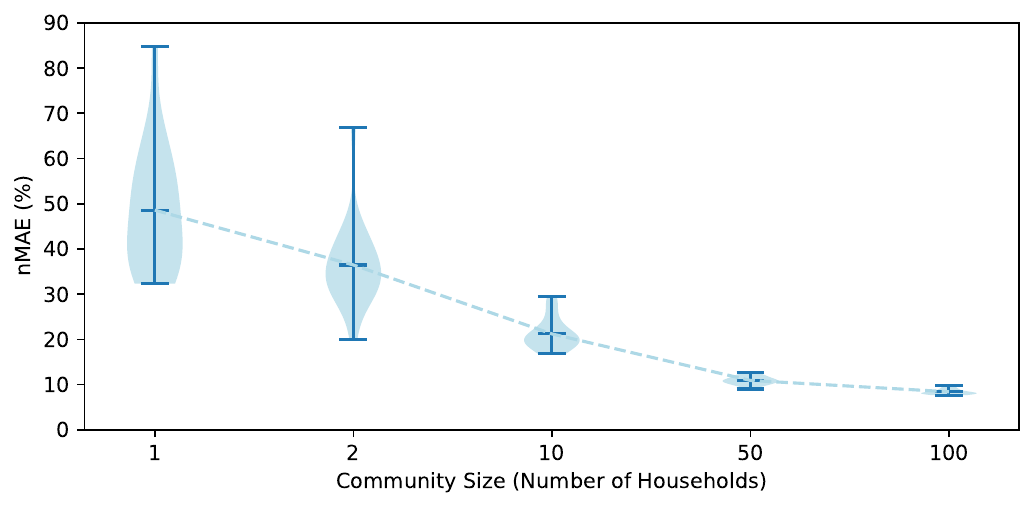}	
    \caption{
    Minimum, arithmetic mean, maximum, and distribution of nMAE across different community sizes, illustrated using the Transformer model. Each setting is based on the results of 20 different profiles, following the baseline configuration.
    }
    \label{Figure_11}
\end{figure}

The impact of the EC aggregation size is depicted in Fig.~\ref{Figure_11} in terms of nMAE. 
The results indicate that nMAE is strongly influenced by community size. In particular, a decrease of nMAE with increasing community size is seen throughout the range investigated. At the same time, the standard deviation decreases with nMAE. The present trend, although illustrated using the Transformer architecture, is consistent across all models in the study (see \ref{AppendixA}).

\begin{figure}[h]
	\centering 
	\includegraphics[width=0.5\columnwidth]{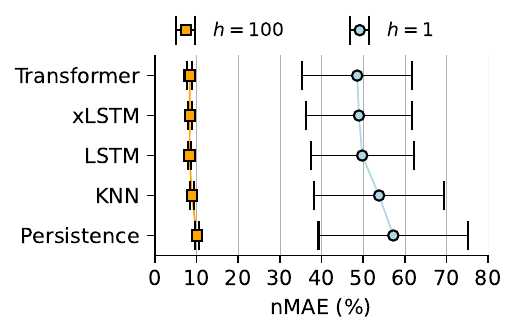}	
	\caption{Forecast nMAE of 20 single households (light blue circles, $h=1$) and 20 communities consisting of 100 households (orange squares, $h=100$). Circles and squares represent the mean values, while the whiskers indicate the standard deviations. All other settings follow the baseline configuration.} 
	\label{Figure_12}
\end{figure}

All forecasting algorithms are evaluated regarding their forecasting capability for different levels of aggregation. The results are depicted in Fig.~\ref{Figure_12} for single households (light blue circles) and for ECs comprised of 100 households (black markers).

Fig.~\ref{Figure_12} demonstrates that the persistence model consistently performs worse than deep learning models at both low and high aggregation sizes. The performance gap gets stronger at higher aggregation levels, such as 100 households, where the forecast deviation is small.

Finally, it is worth emphasizing that all three deep learning models exhibit comparable performance across various community sizes, with only minor differences observed.
As a result, any of these models can be selected for load forecasting.
The model choice can therefore be based on specific needs or preferences, without concerns about significant variations in forecast error. This enables users to prioritize factors such as ease of implementation, computational efficiency, or seamless integration with existing systems.

\subsection{Financial Impact of Forecast Errors}

\begin{figure}
    \centering
    \begin{subfigure}[b]{\textwidth}
        \centering
        \captionsetup{justification=raggedright,singlelinecheck=false}
	\caption{} 
	\includegraphics[width=\columnwidth]{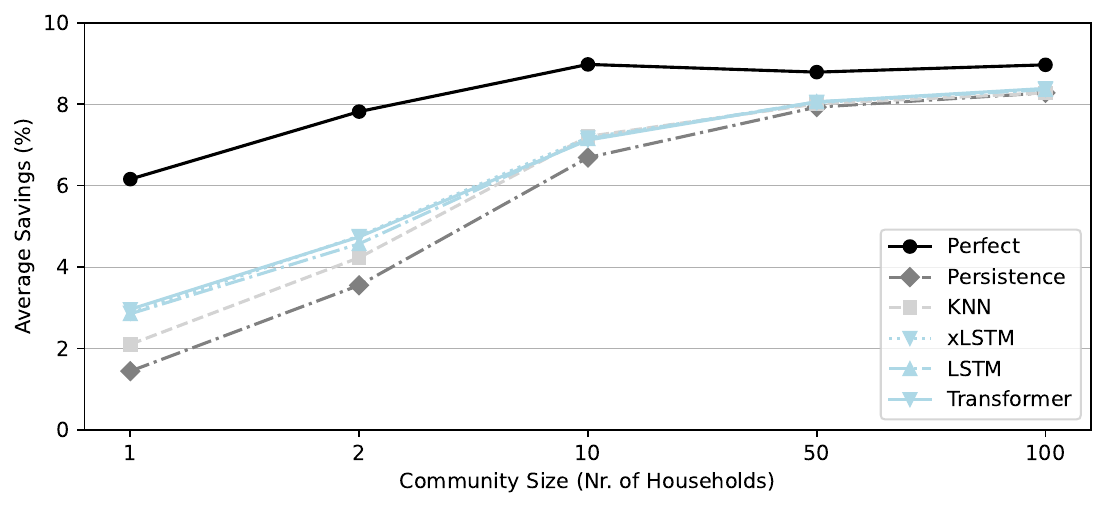}	
    \end{subfigure}
    \begin{subfigure}[b]{\textwidth}
        \centering
        \captionsetup{justification=raggedright,singlelinecheck=false}
        \caption{}
        \includegraphics[width=\textwidth]{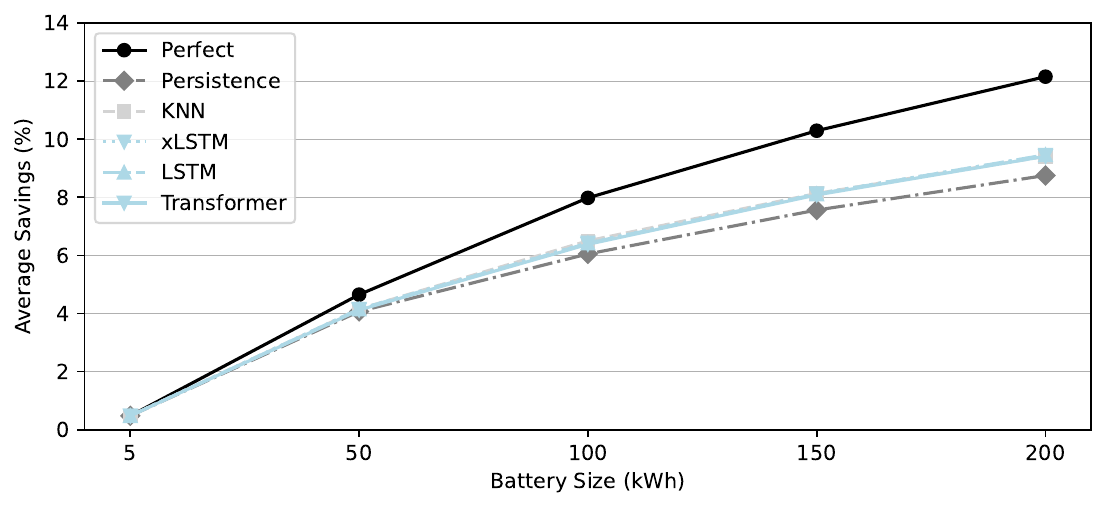}
    \end{subfigure}
    \caption{
    (a) Average relative savings (\%) across different community sizes, with BESS capacity scaled linearly at 12 kWh per household. (b) Average savings (\%) for communities of 10 households across different shared BESS capacities.
    }
    \label{Figure_14}
\end{figure}

To evaluate the average cost savings for the case study, a non-optimized scenario without BESS utilization is computed as the reference case. This reference case corresponds to a use case without any flexibility. For the investigations with utilization of the BESS flexibility, power load predictors (forecasting models) are evaluated against a baseline representing a perfect prediction. The corresponding optimization results are summarized in Fig.~\ref{Figure_14}, with the average savings regarding the size of the EC (a) and the size of the BESS~(b).

Next to the perfect prediction, which assumes perfect knowledge of the load, the best results for a 100-household energy community are achieved using the Transformer leading to cost savings of up to 8.39\%. The remaining two deep learning methods---LSTM and xLSTM---perform comparably well, and the performance differences between the three deep learning methods are minor. 
It is seen that the difference in average savings between the perfect prediction scenario and the various forecasting scenarios is decreasing with an increased level of aggregation. This is attributed to the improved forecast accuracy for larger, aggregated loads. While deep learning models outperform benchmark models especially in smaller ECs, the performance gap narrows as the community size increases. For large ECs (e.g., 100 households), the relative cost savings converge across all forecasting models, ranging from 8.28\% to 8.39\%. However, deep learning models still provide a slightly superior load forecast accuracy, as indicated by lower nMAE values. 

Smaller ECs tend to exhibit higher peak loads relative to their BESS capacity, which limits their ability to shift loads effectively. This limitation is particularly evident under the perfect prediction scenario, where the load shifting potential is constrained not only by the forecast accuracy but also by physical system limitations. In contrast, larger ECs benefit from less volatile load profiles and proportionally larger BESS capacities, enabling more efficient load shifting. Fig.~\ref{Figure_14}(b) illustrates the relationship between average relative savings and BESS size for an EC consisting of 10 households. Relative average savings between $0.5\%$ and $12\%$ are achievable for BESS sizes between $5$ and $200~\mathrm{kWh}$. As the energy demand remains constant but the available flexibility increases with larger BESS sizes, cost savings improve accordingly. It is clear that a $5~\mathrm{kWh}$ BESS is relatively undersized for a 10-household EC and leads to limited savings. 

In this case study, all prediction methods perform relatively well, because for RTP-driven load shifting, the price signal is the most important incentive, which is published one day in advance by EXAA. In contrast, other objectives, such as increasing self-sufficiency or reducing peak load are generally more prediction-sensitive, emphasizing the significance of high-quality predictions~\cite{WOHLGENANNT2024226}. 

\section{Conclusions}
\label{sec:conclusion}

This study critically evaluates the suitability of deep learning models for short-term load forecasting in ECs, comparing them against traditional benchmarks such as persistence prediction and KNN regression. The findings suggest that while deep learning methods are often considered state-of-the-art, their performance is not universally superior---especially in scenarios with limited training data.

To improve the performance of the forecasting methods in data-scarce environments, transfer learning from publicly available synthetic load profiles was applied. In scenarios with limited training data---such as only two months---this method reduced forecasting errors by nearly 2 percentage points. This underscores the versatility and generalization ability of these synthetic profiles for pretraining.

Notably, simple methods like persistence forecasting outperform deep learning models when training data is restricted to six months or less, regardless of the use of transfer learning. However, as the training data size increases, deep learning models show significant improvements in forecast accuracy and are getting less reliant on transfer learning.

In addition to predictive performance, the study quantifies the financial implications of load forecasting accuracy using a mixed-integer linear programming case study. For a community of 10 households with a shared battery energy storage system of $120~\mathrm{kWh}$, the best-performing deep learning model (xLSTM) achieved average energy cost savings of 7.18\%. Interestingly, the simple KNN approach yielded approximately the same cost savings, namely 7.21\%, despite its lower model complexity and forecast accuracy.

Thus, the research questions posed in the introduction of this study can be answered as follows:

\begin{enumerate}
    \item In the context of ECs, deep learning models are not always worth the effort---especially for ECs with less than six months of available data, where simpler models like persistence prediction yield lower forecasting errors.
    \item For applications that nevertheless want to use deep learning methods with only a few month of training data, transfer learning from publicly available synthetic load profiles improves the average forecasting error.
    \item From a financial perspective, simple methods such as KNN can yield cost savings comparable to deep learning models, making them a viable option especially in resource-constrained environments.
\end{enumerate}

In future research, the results provided by this study can be built on by comparing forecasting models across a broader range of optimization scenarios. Nonetheless, this study contributes valuable insights to the design and operation of energy communities and highlights the trade-offs between model complexity, data availability, and practical benefits. These contributions lay a strong foundation for advancing more efficient and practical energy management strategies.

\section*{CRediT authorship contribution statement}

\textbf{Lukas Moosbrugger:} Writing - Original Draft, Visualization, Software, Validation, Methodology, Data curation, Formal analysis, Conceptualization. 
\textbf{Valentin Seiler:} Writing - Review \& Editing, Visualization, Validation, Methodology, Conceptualization.
\textbf{Philipp Wohlgenannt:} Writing - Review \& Editing, Software, Validation, Methodology, Conceptualization. 
\textbf{Sebastian Hegenbart:} Writing - Review \& Editing, Validation, Methodology, Conceptualization.
\textbf{Sashko Ristov:} Writing - Review \& Editing, Validation, Methodology, Conceptualization.
\textbf{Elias Eder:} Writing - Review \& Editing, Visualization, Validation, Methodology, Conceptualization.
\textbf{Peter Kepplinger:} Writing - Review \& Editing, Validation, Conceptualization, Supervision, Resources, Project administration, Funding acquisition.

\section*{Declaration of competing interest}
The authors declare that they have no known competing financial interests or personal relationships that could have appeared to influence the work reported in this paper.

\section*{Data availability}

This paper was created with a strong focus on transparency. Therefore, the data of the prediction method (including smartmeter load, weather, synthetic loads for pretraining, holiday calendar, etc.) is publicly accessible in a GitHub repository~\cite{github_loadforecasting2024}.

Only the EXAA price signal referenced in the case study section cannot be provided due to legal restrictions. However, users are encouraged to obtain this data independently, as it is freely available for download.

\section*{Acknowledgements}
The authors gratefully acknowledge the financial support from the Austrian Research Promotion Agency FFG for the Hub4FlECs project (COIN FFG 898053). The authors also gratefully acknowledge Fiona Feurstein's support in proofreading the manuscript.

\appendix
\section{Extended Results}
\label{AppendixA}

Tables~\ref{table_w_transferlearning} and~\ref{table_wo_transferlearning} present additional results, including those for predictors not shown in the result figures.

\begin{table}
\centering
\scriptsize
\renewcommand{\arraystretch}{1.4} 
\caption{Model comparison across different dimensions using the \textbf{mean nMAE (\%) of 20 load profiles} and 
in brackets \textbf{standard deviation (\%)} of these 20 load profiles. All models in this table use transfer learning from synthetic load profiles.}
\label{table_w_transferlearning}
\begin{tabular}{m{0.78cm} m{1.2cm} m{1.8cm} m{1.8cm} m{1.8cm} m{1.8cm} m{1.8cm} m{1.8cm}}
\hline
\multirow{1}{*}{\textbf{}} & \multirow{1}{*}{\textbf{Setup}} & \textbf{KNN} & \textbf{Persistence} & \textbf{xLSTM} & \textbf{LSTM} & {\textbf{Transformer}} \\
\hline
\multirow{7}{*}{\rotatebox[origin=c]{90}{\textbf{Model Size}}} 
    & 0.1k & \textbf{11.24} (0.93) & 12.79 (1.13) & - (-) & 19.18 (1.82) & 21.39 (2.21) \\ 
    & 0.2k & 11.24 (0.93) & 12.79 (1.13) & - (-) & 11.82 (0.97) & \textbf{10.82} (0.92) \\ 
    & 0.5k & 11.24 (0.93) & 12.79 (1.13) & - (-) & 10.91 (1.00) & \textbf{10.83} (0.83) \\ 
    & 5k & 11.24 (0.93) & 12.79 (1.13) & 11.16 (1.00) & \textbf{10.84} (0.92) & 10.91 (0.91) \\ 
    & 20k & 11.24 (0.93) & 12.79 (1.13) & 11.79 (0.88) & 11.55 (0.95) & \textbf{11.04} (0.97) \\ 
    & 40k & \textbf{11.24} (0.93) & 12.79 (1.13) & 12.60 (1.16) & 11.90 (0.85) & 11.27 (0.99) \\ 
    & 80k & 11.24 (0.93) & 12.79 (1.13) & 12.53 (1.11) & 12.46 (1.23) & \textbf{11.19} (1.10) \\ 
\hline
\multirow{4}{*}{\rotatebox[origin=c]{90}{\shortstack{\textbf{Testset} \\ \textbf{(2013)}}}} 
    & Q1 & 13.70 (1.57) & 14.02 (1.64) & 12.25 (1.66) & 12.48 (1.52) & \textbf{11.73} (1.69) \\ 
    & Q2 & 14.12 (1.31) & 14.54 (1.01) & 12.41 (1.37) & 12.13 (1.07) & \textbf{12.00} (1.21) \\ 
    & Q3 & 11.96 (1.09) & 11.89 (0.97) & 10.40 (0.85) & 10.56 (0.72) & \textbf{10.03} (0.86) \\ 
    & Q4 & 12.59 (1.09) & 12.79 (1.13) & 11.56 (1.36) & 11.56 (1.23) & \textbf{11.20} (1.33) \\ 
\hline
\multirow{5}{*}{\rotatebox[origin=c]{90}{\shortstack{\textbf{Community} \\ \textbf{Size}}}} 
    & 1 & 53.84 (15.54) & 57.25 (17.94) & 49.01 (12.77) & 49.79 (12.37) & \textbf{48.58} (13.16) \\ 
    & 2 & 39.04 (17.77) & 44.98 (17.17) & 36.59 (9.74) & 37.55 (10.63) & \textbf{36.45} (9.01) \\ 
    & 10 & 21.63 (3.43) & 24.86 (3.13) & \textbf{21.12} (3.02) & 21.13 (3.07) & 21.20 (3.39) \\ 
    & 50 & 11.24 (0.93) & 12.79 (1.13) & 11.16 (1.00) & \textbf{10.84} (0.92) & 10.91 (0.91) \\ 
    & 100 & 8.94 (0.40) & 10.10 (0.49) & 8.45 (0.45) & \textbf{8.33} (0.41) & 8.40 (0.61) \\ 
\hline
\multirow{6}{*}{\rotatebox[origin=c]{90}{\textbf{Training Size}}} 
    & 2 mo & 15.81 (1.82) & \textbf{12.79} (1.13) & 18.22 (2.09) & 14.87 (2.54) & 16.56 (2.73) \\ 
    & 4 mo & 15.79 (1.82) & \textbf{12.79} (1.13) & 17.56 (1.83) & 15.24 (1.73) & 16.76 (2.68) \\ 
    & 6 mo & 16.40 (1.79) & \textbf{12.79} (1.13) & 14.57 (1.68) & 13.46 (1.55) & 15.06 (2.13) \\ 
    & 9 mo & 12.59 (1.09) & 12.79 (1.13) & 11.56 (1.36) & 11.56 (1.23) & \textbf{11.20} (1.33) \\ 
    & 12 mo & 11.24 (0.93) & 12.79 (1.13) & 11.16 (1.00) & \textbf{10.84} (0.92) & 10.91 (0.91) \\ 
    & 15 mo & 11.22 (0.92) & 12.79 (1.13) & 10.83 (0.86) & \textbf{10.77} (0.87) & 10.95 (0.91) \\ 
\hline
\end{tabular}
\end{table}

\begin{table*}[tbph]
\centering
\scriptsize
\renewcommand{\arraystretch}{1.4} 
\caption{Model comparison \textbf{without transfer learning} across different dimensions using the \textbf{mean nMAE (\%) of 20 load profiles} and 
in brackets the \textbf{standard deviation (\%)}.
}
\label{table_wo_transferlearning}
\begin{tabular}{m{0.5cm} m{1cm} m{1.8cm} m{1.8cm} m{1.8cm} m{1.8cm} m{1.8cm} m{1.8cm}}
\hline
\multirow{1}{*}{\textbf{}} & \multirow{1}{*}{\textbf{Setup}} & \textbf{KNN} & \textbf{Persistence} & \textbf{xLSTM} & \textbf{LSTM} & \textbf{Transformer} \\
\hline
\multirow{6}{*}{\rotatebox[origin=c]{90}{\textbf{Training Size}}} 
    & 2 mo & 15.81 (1.82) & \textbf{12.79} (1.13) & 19.09 (3.70) & 18.35 (2.70) & 18.10 (3.30) \\ 
    & 4 mo & 15.79 (1.82) & \textbf{12.79} (1.13) & 17.32 (1.92) & 16.34 (2.12) & 16.40 (1.79) \\ 
    & 6 mo & 16.40 (1.79) & \textbf{12.79} (1.13) & 14.83 (1.65) & 14.55 (1.73) & 14.43 (1.54) \\ 
    & 9 mo & 12.59 (1.09) & 12.79 (1.13) & 11.53 (0.96) & 11.60 (1.37) & \textbf{11.35} (1.25) \\ 
    & 12 mo & 11.24 (0.93) & 12.79 (1.13) & 10.94 (0.86) & 10.79 (0.82) & \textbf{10.78} (0.82) \\ 
    & 15 mo & 11.22 (0.92) & 12.79 (1.13) & 10.95 (0.81) & \textbf{10.77} (0.84) & 10.85 (0.83) \\ 
\hline
\end{tabular}
\end{table*}

\begin{table}[htbp]
\centering
\scriptsize
\caption{Optimization results from the case study, comparing different prediction models across varying community sizes. All values represent averages over 20 distinct communities. 
Parameters not explicitly mentioned remain at baseline values, with cost evaluations conducted over Q4 of 2013.
}
\label{tab_case_res}
\begin{tabular}{llrrrrr}
\toprule
Community Size & Model & nMAE (\%) & Costs (\euro{}) & Savings (\euro{}) & Savings (\%) \\
\midrule
1 Household & Unoptimized &  & 299.06 &  &  \\
 & Perfect & 0.00 & 280.65 & 18.41 & 6.16 \\
 & KNN & 53.84 & 292.77 & 6.29 & 2.10 \\
 & Persistence & 57.25 & 294.76 & 4.30 & 1.44 \\
 & xLSTM & 49.01 & 290.49 & 8.57 & 2.87 \\
 & LSTM & 49.79 & 290.54 & 8.52 & 2.85 \\
 & Transformer & 48.58 & 290.21 & 8.85 & 2.96 \\
\midrule
2 Households & Unoptimized &  & 495.71 &  &  \\
 & Perfect & 0.00 & 456.94 & 38.78 & 7.82 \\
 & KNN & 39.04 & 474.73 & 20.98 & 4.23 \\
 & Persistence & 44.98 & 478.09 & 17.62 & 3.55 \\
 & xLSTM & 36.59 & 472.10 & 23.61 & 4.76 \\
 & LSTM & 37.55 & 473.06 & 22.65 & 4.57 \\
 & Transformer & 36.45 & 472.20 & 23.51 & 4.74 \\
\midrule
10 Households & Unoptimized &  & 2218.93 &  &  \\
 & Perfect & 0.00 & 2019.64 & 199.28 & 8.98 \\
 & KNN & 21.63 & 2058.95 & 159.97 & 7.21 \\
 & Persistence & 24.86 & 2070.55 & 148.38 & 6.69 \\
 & xLSTM & 21.12 & 2059.71 & 159.22 & 7.18 \\
 & LSTM & 21.13 & 2060.31 & 158.62 & 7.15 \\
 & Transformer & 21.20 & 2061.00 & 157.93 & 7.12 \\
\midrule
50 Households & Unoptimized &  & 11601.29 &  &  \\
 & Perfect & 0.00 & 10581.28 & 1020.01 & 8.79 \\
 & KNN & 11.24 & 10671.82 & 929.48 & 8.01 \\
 & Persistence & 12.79 & 10681.87 & 919.43 & 7.93 \\
 & xLSTM & 11.16 & 10668.24 & 933.05 & 8.04 \\
 & LSTM & 10.84 & 10666.32 & 934.97 & 8.06 \\
 & Transformer & 10.91 & 10665.71 & 935.58 & 8.06 \\
\midrule
100 Households & Unoptimized &  & 22590.18 &  &  \\
 & Perfect & 0.00 & 20562.84 & 2027.33 & 8.97 \\
 & KNN & 8.94 & 20720.69 & 1869.48 & 8.28 \\
 & Persistence & 10.10 & 20720.22 & 1869.96 & 8.28 \\
 & xLSTM & 8.45 & 20699.30 & 1890.88 & 8.37 \\
 & LSTM & 8.33 & 20701.29 & 1888.89 & 8.36 \\
 & Transformer & 8.40 & 20694.27 & 1895.91 & 8.39 \\
\bottomrule
\end{tabular}
\end{table}

\begin{table}[htbp]
\centering
\scriptsize
\caption{
Case study optimization results for varying battery sizes in 10-household communities. Values are averaged over 20 distinct communities, with all unspecified parameters set to baseline and cost evaluations performed over Q4 of 2013.
}
\label{tab_case_res_2}
\begin{tabular}{llrrrrr}
\toprule
Battery Size & Model & nMAE (\%) & Costs (\euro{}) & Savings (\euro{}) & Savings (\%) \\
\midrule
5 kWh & Unoptimized &  & 2218.93 &  &  \\
 & Perfect & 0.00 & 2208.29 & 10.64 & 0.48 \\
 & KNN & 21.63 & 2208.29 & 10.64 & 0.48 \\
 & Persistence & 24.86 & 2208.29 & 10.64 & 0.48 \\
 & xLSTM & 21.12 & 2208.29 & 10.64 & 0.48 \\
 & LSTM & 21.13 & 2208.29 & 10.64 & 0.48 \\
 & Transformer & 21.20 & 2208.29 & 10.63 & 0.48 \\
\midrule
50 kWh & Unoptimized &  & 2218.93 &  &  \\
 & Perfect & 0.00 & 2115.80 & 103.13 & 4.65 \\
 & KNN & 21.63 & 2126.86 & 92.07 & 4.15 \\
 & Persistence & 24.86 & 2128.54 & 90.39 & 4.07 \\
 & xLSTM & 21.12 & 2126.89 & 92.03 & 4.15 \\
 & LSTM & 21.13 & 2126.87 & 92.06 & 4.15 \\
 & Transformer & 21.20 & 2127.52 & 91.40 & 4.12 \\
\midrule
100 kWh & Unoptimized &  & 2218.93 &  &  \\
 & Perfect & 0.00 & 2041.77 & 177.15 & 7.98 \\
 & KNN & 21.63 & 2074.64 & 144.28 & 6.50 \\
 & Persistence & 24.86 & 2084.62 & 134.31 & 6.05 \\
 & xLSTM & 21.12 & 2076.16 & 142.77 & 6.43 \\
 & LSTM & 21.13 & 2076.56 & 142.37 & 6.42 \\
 & Transformer & 21.20 & 2077.32 & 141.61 & 6.38 \\
\midrule
150 kWh & Unoptimized &  & 2218.93 &  &  \\
 & Perfect & 0.00 & 1990.58 & 228.35 & 10.29 \\
 & KNN & 21.63 & 2038.51 & 180.42 & 8.13 \\
 & Persistence & 24.86 & 2051.17 & 167.75 & 7.56 \\
 & xLSTM & 21.12 & 2038.71 & 180.22 & 8.12 \\
 & LSTM & 21.13 & 2038.95 & 179.98 & 8.11 \\
 & Transformer & 21.20 & 2039.73 & 179.20 & 8.08 \\
\midrule
200 kWh & Unoptimized &  & 2218.93 &  &  \\
 & Perfect & 0.00 & 1949.39 & 269.53 & 12.15 \\
 & KNN & 21.63 & 2010.08 & 208.84 & 9.41 \\
 & Persistence & 24.86 & 2024.78 & 194.15 & 8.75 \\
 & xLSTM & 21.12 & 2009.31 & 209.62 & 9.45 \\
 & LSTM & 21.13 & 2009.07 & 209.86 & 9.46 \\
 & Transformer & 21.20 & 2010.11 & 208.82 & 9.41 \\
\bottomrule
\end{tabular}
\end{table}

\section{Impact of the Testing Quarter of the Year}
\label{AppendixB}

In practice, time series forecasting relies on historical training data to forecast future time horizons. 
Using future data as training data is considered unrealistic, as future knowledge might be leaked to the past. Unfortunately, based on the available training corpus, it is not feasible to maintain 12 months of training data prior to each potential test set period throughout the year. To address this, the year 2013 is divided into four quarters, with each quarter sequentially used as the test set while the remaining three quarters form the training set.
This seasonal-data-split strategy maximizes the utility of the dataset and provides a robust framework for evaluating model performance across the entire year. 

The results of this experiment (presented in Tab.~\ref{table_w_transferlearning}) demonstrate that the load forecasting models employed in this study perform consistently well throughout the entire year.
All quarters show consistent results with similar error ranges, confirming that Q4 2013 serves as a representative and balanced benchmark for evaluation.

\section{Model Configuration}
\label{AppendixC}

Table \ref{tab:model_comparison2} lists all the configurations of the deep learning models considered.

\begin{table}
\centering
\scriptsize
\renewcommand{\arraystretch}{1.25} 
\caption{Configuration of the xLSTM, LSTM, and Transformer of all model sizes. Note that the reported parameter counts are approximate, as slight variations arise due to architectural differences between models. }
\label{tab:model_comparison2}
\begin{tabular}{m{1.8cm} m{1.8cm} m{1.9cm} m{3.1cm} m{2.9cm}}
\hline
\textbf{} & \textbf{Model Size} & \textbf{xLSTM} & \textbf{LSTM} & \textbf{Transformer} \\ \hline
\textbf{Input} & & \multicolumn{3}{c}{Input Shape = (Batchsize, 24 Hours, 20 Features)} \\ \hline
\textbf{Projection} & & Yes & No & Yes \\ \hline
\textbf{Positional Encoding} & & Yes & No & Yes \\ \hline
\textbf{Sequence Model}  & \textbf{0.1k} & 
\begin{tabular}[t]{@{}l@{}}Blocks = 1 \\ Heads = 1 \\ Features = 1\\ sLSTM at 0\end{tabular} &
\begin{tabular}[t]{@{}l@{}}Layer 1 = 1 LSTM \\ Layer 2 = 1 LSTM\\Layer 3 = Dense(4)\\Layer 4 = Dense(4)\end{tabular} &
\begin{tabular}[t]{@{}l@{}}Layers = 1 \\ Heads = 2 \\ Feedforward Dim = 5 \\ Features = 2\end{tabular} \\ \cline{2-5}
& \textbf{0.2k} & 
\begin{tabular}[t]{@{}l@{}}Blocks = 1 \\ Heads = 1 \\ Features = 1 \\ sLSTM at 0\end{tabular} &
\begin{tabular}[t]{@{}l@{}}Layer 1 = 1 Bi-LSTM \\ Layer 2 = 1 Bi-LSTM\\Layer 3 = Dense(4)\\Layer 4 = Dense(4)\end{tabular} &
\begin{tabular}[t]{@{}l@{}}Layers = 1 \\ Heads = 2 \\ Feedforward Dim = 5 \\ Features = 4\end{tabular} \\ \cline{2-5}
& \textbf{0.5k} & 
\begin{tabular}[t]{@{}l@{}}Blocks = 1 \\ Heads = 2 \\ Features = 2 \\ sLSTM at 0\end{tabular} &
\begin{tabular}[t]{@{}l@{}}Layer 1 = 2 Bi-LSTM \\ Layer 2 = 2 Bi-LSTM\\Layer 3 = Dense(5)\\Layer 4 = Dense(5)\end{tabular} &
\begin{tabular}[t]{@{}l@{}}Layers = 1 \\ Heads = 2 \\ Feedforward Dim = 6 \\ Features = 6\end{tabular} \\ \cline{2-5}
& \textbf{5k} & 
\begin{tabular}[t]{@{}l@{}}Blocks = 2 \\ Heads = 4 \\ Features = 8\\ sLSTM at 1\end{tabular} &
\begin{tabular}[t]{@{}l@{}}Layer 1 = 8 Bi-LSTM \\ Layer 2 = 9 Bi-LSTM\\Layer 3 = Dense(30)\\Layer 4 = Dense(20)\end{tabular} &
\begin{tabular}[t]{@{}l@{}}Layers = 1 \\ Heads = 4 \\ Feedforward Dim = 90 \\ Features = 20\end{tabular} \\ \cline{2-5}
& \textbf{20k} & 
\begin{tabular}[t]{@{}l@{}}Blocks = 2 \\ Heads = 4 \\ Features = 32\\ sLSTM at 1\end{tabular} &
\begin{tabular}[t]{@{}l@{}}Layer 1 = 22 Bi-LSTM \\ Layer 2 = 20 Bi-LSTM\\Layer 3 = Dense(30)\\Layer 4 = Dense(20)\end{tabular} &
\begin{tabular}[t]{@{}l@{}}Layers = 1 \\ Heads = 4 \\ Feedforward Dim = 400 \\ Features = 20\end{tabular} \\ \cline{2-5}
& \textbf{40k} & 
\begin{tabular}[t]{@{}l@{}}Blocks = 4 \\ Heads = 4 \\ Features = 32\\ sLSTM at 1\end{tabular}&
\begin{tabular}[t]{@{}l@{}}Layer 1 = 42 Bi-LSTM \\ Layer 2 = 20 Bi-LSTM\\Layer 3 = Dense(30)\\Layer 4 = Dense(20)\end{tabular} &
\begin{tabular}[t]{@{}l@{}}Layers = 1 \\ Heads = 4 \\ Feedforward Dim = 400 \\ Features = 40\end{tabular} \\ \cline{2-5}
& \textbf{80k} & 
\begin{tabular}[t]{@{}l@{}}Blocks = 4 \\ Heads = 8 \\ Features = 40\\ sLSTM at 1\end{tabular} &
\begin{tabular}[t]{@{}l@{}}Layer 1 = 70 Bi-LSTM \\ Layer 2 = 21 Bi-LSTM\\Layer 3 = Dense(30)\\Layer 4 = Dense(20)\end{tabular} &
\begin{tabular}[t]{@{}l@{}}Layers = 2 \\ Heads = 8 \\ Feedforward Dim = 400 \\ Features = 40\end{tabular} \\ \hline
\textbf{Output Layer} & & \multicolumn{3}{c}{1 Neuron, No Activation} \\ \hline
\textbf{Output} & & \multicolumn{3}{c}{Output Shape = (Batchsize, 24 Hours, 1 Powervalue)} \\ \hline
\end{tabular}
\end{table}

\bibliographystyle{elsarticle-num}
\bibliography{references}

\end{document}